\RequirePackage{fix-cm}
\pdfoutput=1
\documentclass{svjour3}                     

\smartqed  

\usepackage{graphicx}
\usepackage{natbib}
\usepackage{subfigure}
\usepackage{amsmath}

\begin{document}

\title{Feature base fusion for splicing forgery detection based on neuro fuzzy}

\author{Habib Ghaffari Hadigheh         \and
        Ghazali bin sulong 
}

\institute{Habib Ghaffari-Hadigheh \at
              Universiti Tecknologi Malaysia (UTM) \\
              Tel.: +98-914-3091405\\
              \email{ghhabib2@live.utm.my}           
           \and
           Ghazali bin sulong \at
              Universiti Malaysia Terengganu (UMT) \\
              Tel.: +60-177-467128\\
              \email{ghazali.s@umt.edu.my}
}

\date{Received: date / Accepted: date}

\maketitle

\begin{abstract}

Most of researches on image forensics  have been mainly focused on detection of artifacts introduced by a single processing tool. They  lead in the development of many specialized algorithms looking for one or more particular footprints under specific settings. Naturally, the performance of such algorithms are not perfect, and accordingly the provided output  might be noisy, inaccurate and only partially correct. Furthermore, a forged image  in practical scenarios is often the result of utilizing several tools  available by  image-processing software systems. Therefore, reliable tamper detection requires developing more poweful tools to deal with various tempering scenarios. Fusion of forgery detection tools based on Fuzzy Inference System has been used before for addressing this problem. Adjusting the membership functions and defining proper fuzzy rules for attaining to better results are time-consuming processes. This can be accounted as main disadvantage of fuzzy inference systems. In this paper, a Neuro-Fuzzy inference system for fusion of forgery detection tools is developed. The neural network characteristic of these systems provides appropriate tool for automatically adjusting  the membership functions. Moreover, initial fuzzy inference system is generated based on fuzzy clustering techniques. The proposed framework is implemented and validated on a benchmark image splicing data set in which three forgery detection tools are fused based on adaptive Neuro-Fuzzy inference system. The  outcome of the proposed method reveals that applying Neuro Fuzzy inference systems could be a better approach for fusion of forgery detection tools.
\keywords{Machine Learning \and Splicing Forgery Detection \and Image Processing.}
\end{abstract}

\section{Introduction}
\label{intro}

Nowadays, exploitation of digital images and photos taken by various smart recording devices is getting more prevalent along with the extensive use of social media like Facebook, Twitter, and Instagram. Individuals tend to spend a considerable amount of time surfing the Internet, and share almost all moments of their lives, as well as events in their districts such as festivals, concerts or some dreadful incidents e.g. terrorist attacks, armed robberies or anti-human rights behaviors. Recently, so many devices exist for producing digital images, and almost each communication device has access to the Internet and is equipped with a digital camera which produces very high-resolution pictures. These snapshots and professionally taken images are of very big advantage, since they are accounted as an inevitable part of the accompanying contexts in the media, and they help to elucidate different aspects of events, while at the same time they could be a source of troublemaking consequences both in local and global scales.

Images might be the mean of deliberately cynical distortion of a reality, and aimed to convey special messages without other corroborating documents. A manipulated image taken from a protest may be intended to exaggerate the crowd thronged in supporting of an idea which might be ignored by the authorities. Distorted shots that are claimed have been taken from private moments of celebrities might be used for extortion or imposing them to act in favor of third parties which are the anomaly in their background. This innate and potential quality of pictures in general and digital image in particular, would make them susceptible for forgery by   digital image editing and post processing.

Above mentioned issues assert the necessity for providing reliable tools to authenticate the originality of such images, which attracted many researchers to work on detecting the possible traces of forgery. Forgery detection methods are categorized in two general branches, the active and the passive methods~[\cite{farid2009exposing}]. The former is based on the idea of inserting information inside the images and use them for authentication and proving of their integrity. This information is usually inserted either at the time of taking the photos or during the post-processing operations. This approach needs special kind of hardware while the information could easily be altered~[\cite{katzenbeisser2000information,cox2003digital}]. The latter exploits the image data to detect the possible clues of forgery.  Since most of  devices in the market are not equipped with such complicated and fancy tools to accomplish the first method, and consequently taken photos are free of such embedded information, passive  methods are more practical~[\cite{farid2009exposing}].

There are different kinds of passive methods for detecting possible forgeries, while none of them claims to have 100 percent accuracy, since forgers sometimes professionally manipulate images such that detection of forgery becomes very hard and even an impossible job~[\cite{kirchner2007tamper,kirchner2008hiding}]. Image splicing, copy-paste attack, and image retouching are three examples of passive methods. Most of the researches just focus on one type of these methods, because they usually don't use common features of photos during their process~[\cite{avcibas2004classifier,hsiao2005detecting}].

In this research, we only focus on  image splicing methods. These methods are applicable when the forged image are produced by integration of more than one image. Splicing detection is the most challenging problem while there is almost no ultimate guaranteeing solution~[\cite{Rasse2014}]. To overcome this uncertainty, we use Neuro-Fuzzy Inference System~(NFIS) for the fusion of features extracted by three different splicing forgery detection methods; Discrete Wavelet Transformation (DWT) decompression [\cite{fu2006detection}], Edge images based on Gray Level Co-occurrence Matrix~(GLCM)~[\cite{wang2009effective}], and N-Run Length~[\cite{dong2009run}]. To the best of our knowledge this is the first study on using NFIS in the combination of these methods.

The reminder of this paper is organized as follows: In Section~\ref{sec:1} a literature of problem is reviewed. The methodology of the study is described in Section~\ref{sec:2}. Section~\ref{sec:3} depicts experimental results, and the final section includes the discussion on the results and potential future work directions.

\section{Problem Background}
\label{sec:1}
Image splicing is a kind of forgery attack implemented by integration of some  fragments from the same image or others, without further post-processing such as smoothing of boundaries among adjacent fragments~[\cite{zhang2008study}]. Several researches have been conducted on splicing forgery detection. One group focused on detecting the possible forgery using statistical analysis of the image pixels information, others diverted their focus on deriving the inconsistency of light directions, to discover the potentially existent splicing forgery traces~[\cite{redi2011digital}].

Any splicing operation, even when the forger tries to mask it with blending techniques, leaves some traces in the image statistics. Subsequently, these statistical inconsistencies could be used for locating the manipulated place . The primary application of this approach traces back to forgery detection  in digital signals~[\cite{farid1999detecting}]. The main idea was that, when digital data are deformed, spectrum analysis could be used as a detection mean. This tool had been adapted for 2d signal in image processing~[\cite{Ng2004}]. For more details we refer the interested readers to~[\cite{birajdar2013digital}].

Illumination analysis is another trend for splicing forgery detection where illumination directions could be detected by processing the intensity of the colors in the neighboring pixels. The majority of objects in an authenticated image are supposed to have consistent light directions~[\cite{redi2011digital}]. Even though,  modern editing tools allow concealing the traces of splicing in a deceptive manner, but it is not always possible to match the lighting conditions of the fragments even when it has been perfumed by professionals. One of the pioneering attempts  aimed to estimate the incident light direction of different objects in order to highlight the possible mismatches~[\cite{Johnson2005}]. Further findings are reported in~[\cite{johnson2007exposingeye,Zhang2009}].

In spite of numerous practical findings in this area, all of the proposed methods suffer from a tangible degree of indeterminacy. This imperfection due to performing post-processing operations for hiding the traces of forgery or utilizing lossy compression formats. This problem is referred to  as “Uncertainty’’, and could be addressed by diverse approaches. One is to fuse more than one detection tool, either extracted features by each tool are intergraded before decision making~[\cite{he2006detecting,dong2009run,chetty2010nonintrusive}],or each tool individually results to a decision and a combination rule produces a final decision~[\cite{barni2012fuzzy}].

Using fuzzy as a combination rule in the fusion of forgery detection tools had been overlooked until it has been studied in~\cite{chetty2010nonintrusive,barni2012fuzzy}]. The main disadvantage of their method is that, adjusting membership functions in order to acquire accurate results is a time-consuming process. In this paper, we develop the idea presented in [\cite{barni2012fuzzy}], and introduce a methodology in order to detect splicing forgery automatically using Neuro-Fuzzy fusion. The next section describes the steps of the proposed methodology.

\section{Proposed Methodology}
\label{sec:2}
\begin{figure}
\centering
\includegraphics[width=1\textwidth]{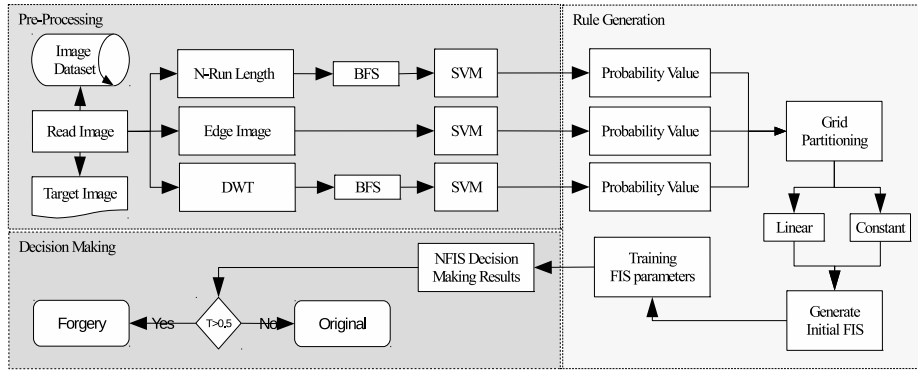}
\caption{Methodology used for fusion of forgery detection tools}
\label{fig:1}       
\end{figure}

The suggested method aims to combine three existing tools; DWT decompression, Edge images based on GLCM, and N-Run Length. In the sequel, three main steps are described in details. Preprocessing is the first step that has the responsibility to extract necessary data for the subsequent steps. This data includes features vectors from images for each tool and they are converted to  decision values by a Support Vector Machine (SVM).  These decision values are used for fusion based on an NFIS in the second step,. Analysis of the results and making final decisions are carried out in the last step. Fig.~\ref{fig:1} illustrates the framework for fusion of forgery detection tools.

\subsection{Preprocessing Step}
\label{sec:2:1}
\begin{figure}
\centering
\includegraphics[width=1\textwidth]{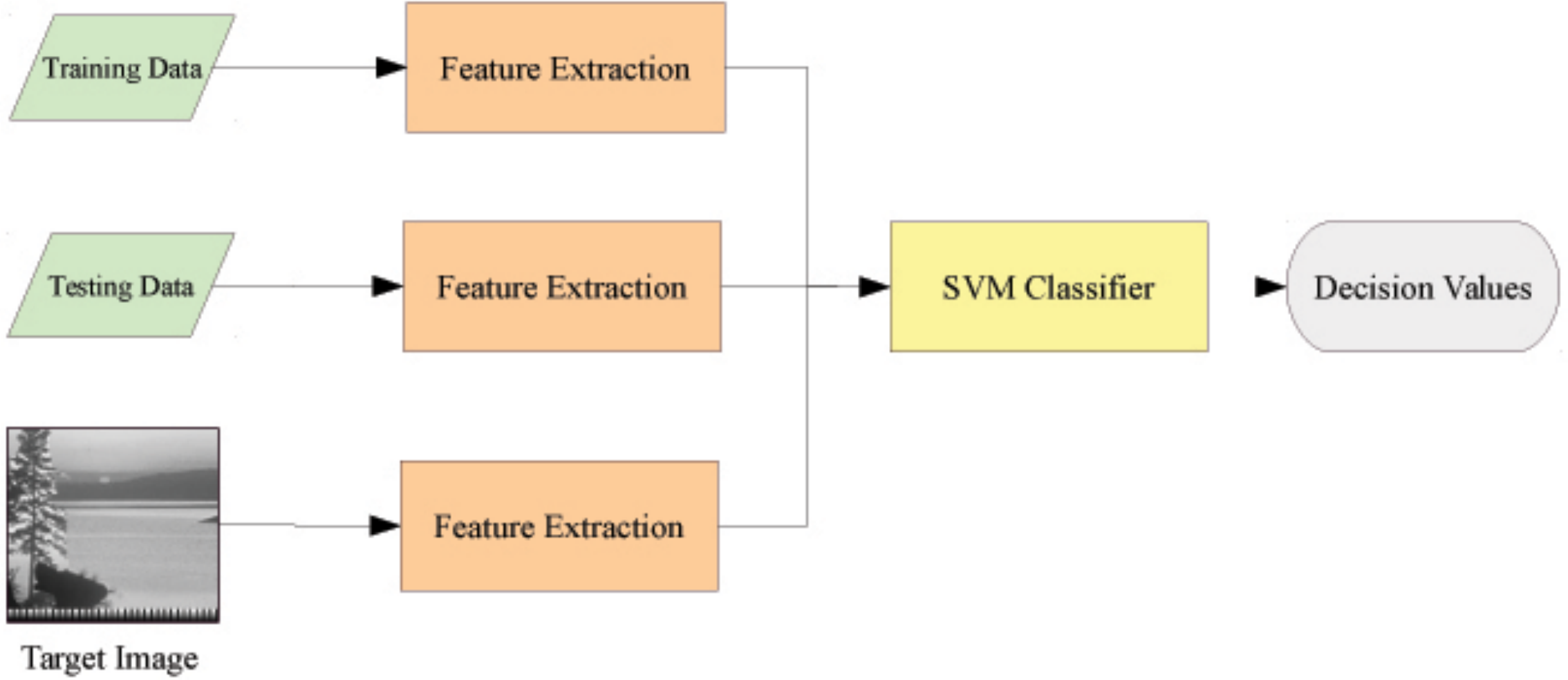}
\caption{Process of training/testing for SVM}
\label{fig:3}
\end{figure}

The first step is to make decision based on each forgery detection tool. Extraction of the features is the most important part of this step. The size of feature vectors in the Edge Image based on GLCM is reasonable~[\cite{wang2009effective}], while Boosting Feature Selection (BFS) algorithm is used to reduce the size of feature vectors in DWT decompression~[\cite{fu2006detection}] and N-Run Length~[\cite{dong2009run}].
Using BFS with an acceptable number of iterations, makes it possible to generate effective feature vectors with the smaller dimensions, an as a result, the main classification step could be accomplished faster. AdaBoost based feature selection system preserves a probability distribution over the training samples. These probabilities are assumed identical at the beginning of the learning stage. Adaboost adjusts them on a series of cycles via a weak learning algorithm. For each training sample, it affiliates a weight and these weights are updated by a multiplicative rule based on the errors of the former learning step. This process is carried out by giving the priority to those samples that are not classified correctly by the previous learning weak classifier. Thus, the samples with lower errors during the weak learning process have greater weights~[\cite{majid2008face}]. Subsequently, a BFS system based on basic BFS introduced in~[\citep{tieu2004boosting}] is designed. Finally, SVM classifier with Radial Bases Function (RBF) kernel is used for stable classification, which provides a decision value of each tool for each individual image. Fig.~\ref{fig:3} depicts the process of generating decision values based on SVM classifier.

\subsection{Rule Generation and Fusion Step}
\label{sec:2:2}
\begin{figure}
\centering
{
\includegraphics[width=0.75\textwidth]{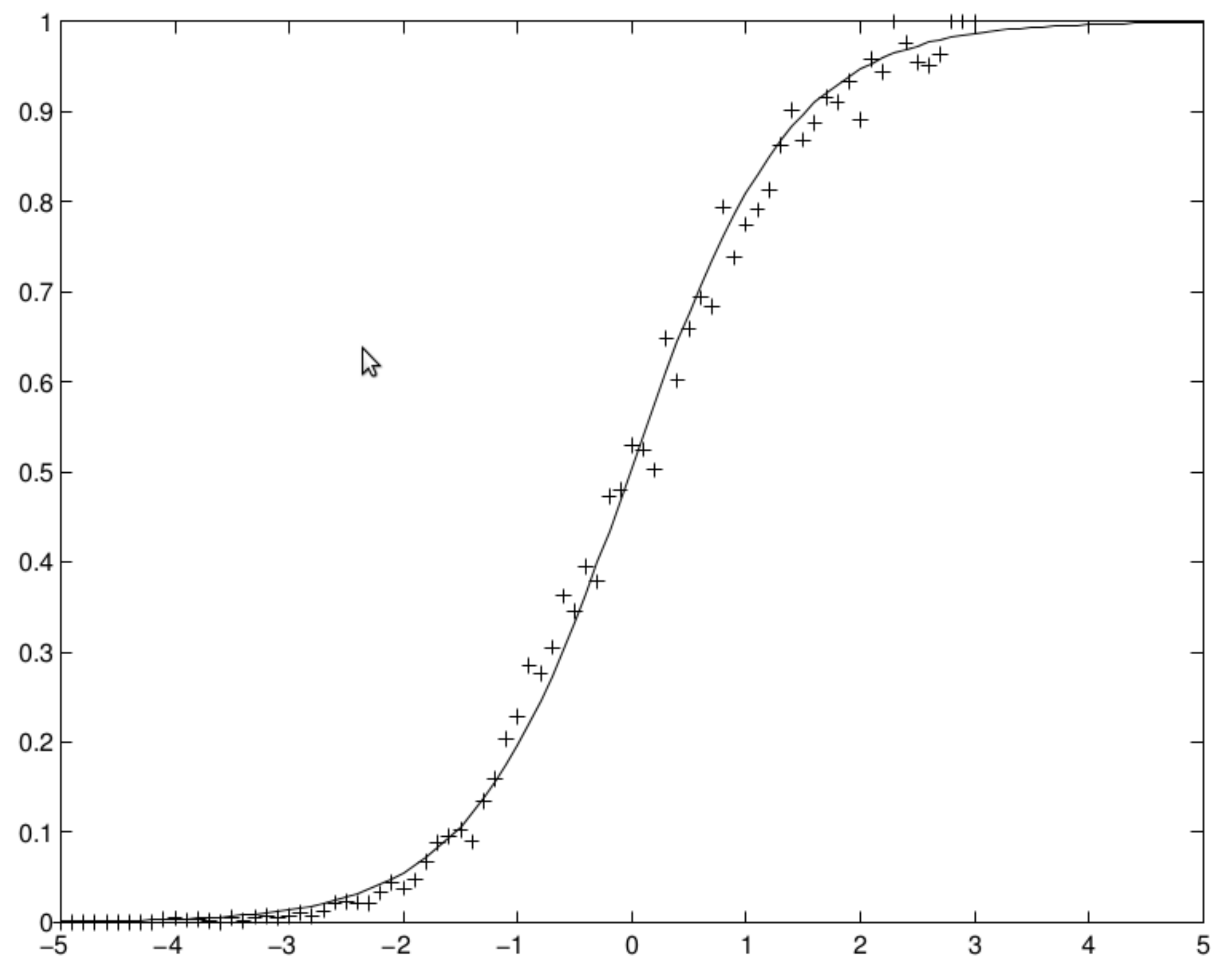}
}
\caption{Sample of sigmoidal base function curve, generated based on an SVM decision values.}
\label{fig:4}
\end{figure}

Let us discuss the process of converting the decision values generated in the previous step into input values for fusion of forgery detection tools. All  decision values are denoted as the detection rate, and should be standardized as a requirement for an NFIS. For this purpose, we have to convert them to numbers in $[0,1]$~[\cite{platt1999probabilistic}]. The main idea is to extract the probability $P(class/input)$ from the decision values of the SVM classifier. To this end, a sigmoid-based function is considered that fits with the output of the classifier to find two parameters $A$ and $B$ of Eq.~\ref{eqn:1}.
\begin{equation}\label{eqn:1}
p(f)=\displaystyle\frac{1}{1+\exp(Af+B)},
\end{equation}
where $f$ is the decision value of the tool. Fig.~\ref{fig:4} depicts a sample of the generated sigmoid function. The plus signs stand for the converted decision values which are fitted to an appropriate sigmoidal function.

\begin{figure}
\centering
{
\includegraphics[width=1\textwidth]{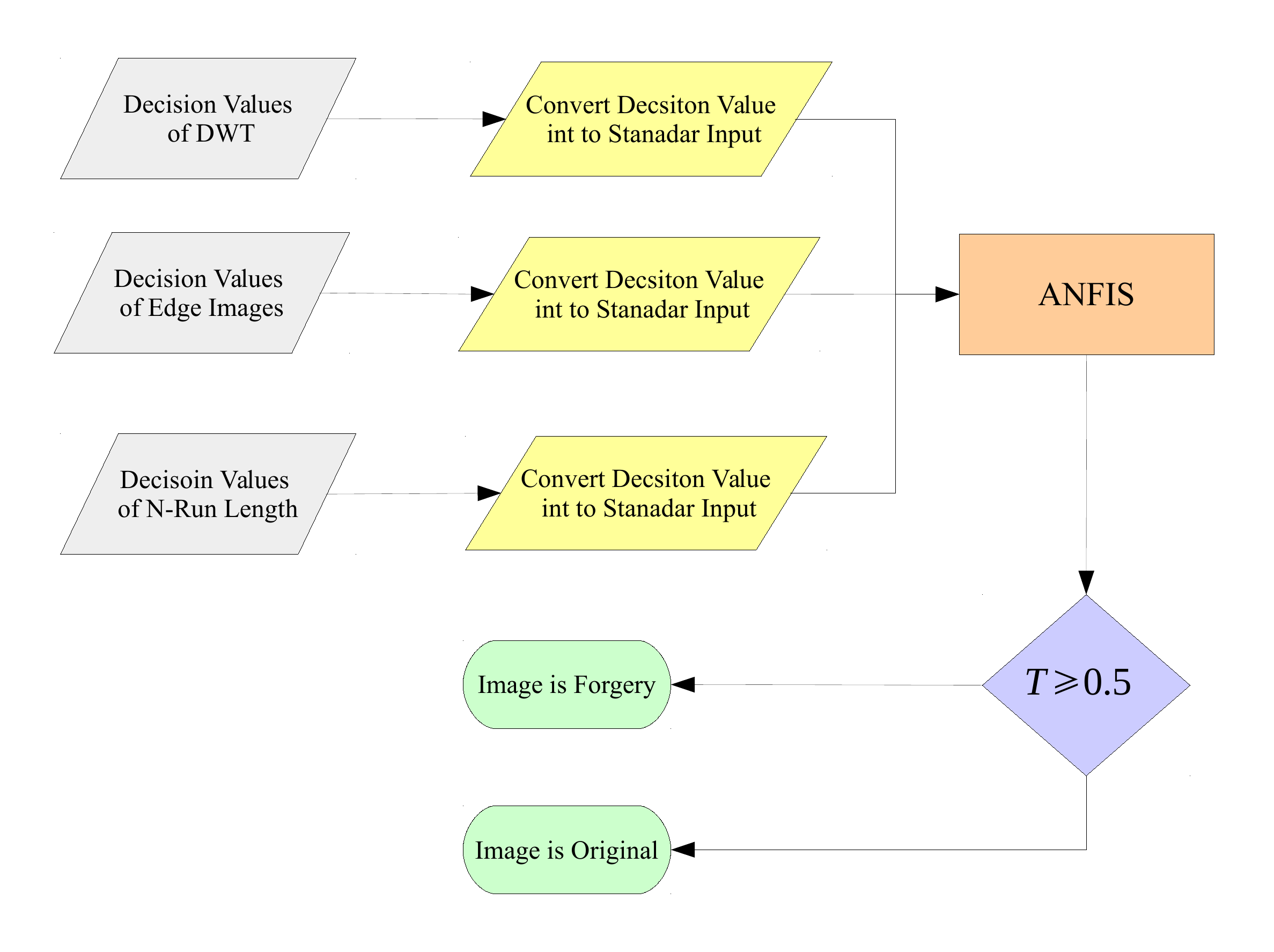}
}
\caption{ANFIS training/testing process}
\label{fig:5}
\end{figure}

Now, these scaled values are considered as inputs for training/testing of the NFIS system, which integrates these values as a single decision value. Implementing details are presented in Section~\ref{sec:3}, and schematically depicted in Fig.~\ref{fig:5}.

\subsection{Evaluation and Comparison}
\label{sec:2:3}

Sensitivity and specificity are used for the construction of the final results. Let us define the following notations.
\begin{itemize}
  \item \textbf{True positive (TP).} The image is detected as a forged one, and it is really a forged image.
  \item \textbf{False Positive (FP).} The image is detected as an authenticated one, while it is a forged image.
  \item \textbf{True Negative (TN).} The image is detected as an authenticated one, and it is a real authenticate image.
  \item \textbf{False Negative (FN).} The image is detected as forged one, while it is an authenticated image.
\end{itemize}

Here, the sensitivity of the test stands for the proportion of images identified to be forged. Mathematically,
\begin{align*}
\text{sensitivity} & = \frac{\text{number of TP}}{\text{number of TP} + \text{number of FN}}, \\ \\& = \frac{\text{number of TP}}{\text{total number of forged image in test samples}}. \\
\end{align*}
Further, specificity is the proportion of authenticated images identified not to be forged. In mathematical notion, it is calculated as
 \begin{align*}
 \text{specificity} & = \frac{\text{number of TN}}{\text{number of TN} + \text{number of FP}}, \\ \\ & = \frac{\text{number of TN}}{\text{total number of authenticate images in test samples}}.
 \end{align*}

\section{Experimental Results}
\label{sec:3}

\begin{figure}
\centering
\subfigure[]{\label{fig:2:a}\includegraphics[width=0.4\textwidth]{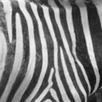}}
\subfigure[]{\label{fig:2:b}\includegraphics[width=0.4\textwidth]{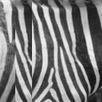}}

  \caption{Image (a) is  an original one and,  (b) is the same  which is spliced using a fragment of another original image.}
  \label{fig:2}
\end{figure}

The proposed methodology was implemented on the benchmark data set \emph{Splicing Forgery Detection Data Set}, provided by the Columbia University's Computer Graphics and User Interfaces Lab~[\cite{DVMMDataset}]. It consists of 933 authenticated and 912 spliced image blocks of the size of $128\times128$. All images are in gray scale mode, and their blocks have been extracted from the CalPhotos image set. The splicing operation was down by cutting some part of the original image blocks and pasting inside the other original one, without any post processing operation. Fig.~\ref{fig:2} depicts two images of this data set, one is authenticated and the other is its forged version.
LIBSVM; a library for support vector machine has been utilized for implementation of the SVM classifier in Matlab~[\cite{chang2011libsvm}]. Its RBF kernel has two  parameters $C$ and $\gamma$, must be specified before starting the training phase. Empirical results denote that RBF with default parameters values may imply to  misclassification or overfitting of the classifier. Therefore, a method based on grid search was proposed and implemented [\cite{dong2009run}].

All the calculations and implementations were performed using a desktop computer with an Intel(R) Core(TM)i7-2630QM CPU, 2.00 $\times$ 8 GHz processor, and 8 Gb of RAM on the Ubuntu 14.04 LTS operating system platform. Image processing toolbox of Matlab 8.1 is used for converting images to the matrices of the size $128\times128$. For obtaining the results, we carried out five runs,  each run includes all  images of the data set. In each run, 90\% of images were randomly selected for training and the rest were left for testing of both SVMs and NFIS. For generating scaled values from each SVM classifier result, a method based on the one introduced in [\cite{platt1999probabilistic}] was used. The image is considered the most authenticated as the associated output value approaches to one. Accordingly, authenticated images of the data set were labeled 1, and the others 0 in both training and testing steps. Complete algorithm and details on the performance of this method are presented thoroughly in [\cite{platt1999probabilistic}], and a Matlab implementation is developed by [\cite{lin2007note}].

We examined different kinds of Neuro-Fuzzy system to find out which of them has better accuracy rate. The input membership functions are considered as Gaussian. However, output membership function is selected either constant or linear, depending on the limitation of the fuzzy toolbox of Matlab. For implementation of NFIS, we applied the Matlab Fuzzy Toolbox. Based on the limitation of NFIS in Matlab, we are just able to implement Adaptive Neuro Fussy Inference System (ANFIS)~[\cite{fzlmtb,4403084}]. We classified the final results based on the threshold used by [\cite{barni2012fuzzy}], when the final decision value of NFIS is greater than $0.5$, the image was considered as authenticated one, and otherwise as a forged image. Fig. ~\ref{fig:5} depicts the flowchart of the evaluation step as well.

The results are reported in Tables~\ref{tab:1} and~\ref{tab:2}. The former depicts the best obtained results of sensitivity measurement, and the latter shows the results of specificity. Five columns are included in both tables; the first column denotes the number of selected features by BFS, which varies from 30 to 100.  The case when all all features are considered stands at the last row. The subsequent three columns  respectively denote the results of DWT, Edge Rune base on GLCM and N-Run Length, and the final column reports the results of the fusion based on NFIS.

As shown in Table~\ref{tab:1}, the best sensitivity for DWT and N-Run Length is achieved when all of the features are used, while this is not the case in others. The accuracy rate of sensitivity for NFIS reveals that there is no necessity for considering all features. Moreover, the results of the NFIS is as promising as the others. As denoted in Table~\ref{tab:2}, the behavior of specificity is contrariwise, the best results are obtained for Edge Rune base on GLCM and NFIS when all features are applied. However, the NFIS competes with the Edge Rune base on GLCM in specificity detection and approximately better than the two others.

\begin{table}

\caption{Results of Fusion in terms of Sensitivity}
\label{tab:1} 
\begin{tabular}{lllll}
\hline\noalign{\smallskip}
Features  Number & DWT & Edge Images & N-Run Length & NFIS  \\
\noalign{\smallskip}\hline\noalign{\smallskip}
30 & 82.32\% & 50.61\% & 59.76\% & 85.37\% \\
50 & 80.00\% & 50.30\% & 55.76\% & 83.03\% \\
75 & 83.54\% & 51.22\% & 59.76\% & 86.59\% \\
100 & 82.32\% & 50.00\% & 59.76\% &  82.93\% \\
All & 89.47\% & 32.26\% & 67.76\% & 72.37\% \\
\noalign{\smallskip}\hline
\end{tabular}
\end{table}

\begin{table}

\caption{Results of Fusion in terms of Specificity}
\label{tab:2}      
\begin{tabular}{lllll}
\hline\noalign{\smallskip}
Features  Number & DWT & Edge Images & N-Run Length & NFIS  \\
\noalign{\smallskip}\hline\noalign{\smallskip}
30 & 72.54\% & 74.19\% & 67.08\% & 71.61\% \\
50 & 70.55\% & 74.85\% & 56.44\% & 85.28\% \\
75 & 76.13\% & 79.73\% & 63.23\% & 70.32\% \\
100 & 76.77\% & 80.65\% & 63.23\% &  81.29\% \\
All & 74.00\% & 86.67\% & 66.67\% & 91.33\% \\
\noalign{\smallskip}\hline
\end{tabular}
\end{table}

 \section{Conclusion}
 \label{sec:4}

 In this paper, we presented a methodology for fusion of features on the decision level, with the aim of  automatically finding  possible forgery traces   in grayscale images. For this propose, we integrated the scaled decision values of three different tools using the NFIS. To achieve lower run time, we applied a BFS algorithm for detecting more efficients among all features.

 Based on the experimental results, our proposed method has almost promising results in comparison with individual tools. It seems that considering a trade-off between two objectives sensitivity and specificity is necessary for further investigation. Based on the limitation of our benchmark data set, we experimented only on  grayscale images, while adapting the methodology on color images could be challenging. We restate that reducing the run time was not our main concern. Nevertheless, we observed that decreasing  the number of features has the positive effect on the run time. Further study can be conducted using recently advanced machine learning methods such as deep learning.

\begin{acknowledgements}
We sincerely appreciate Columbia University Graphic Lab for the help in providing splicing forgery detection Data-set. We also acknowledge the UTM for providing facilities and fretful environment for this research.
\end{acknowledgements}

\bibliographystyle{spbasic}
\bibliography{reference}

\begin{thebibliography}{30}
\providecommand{\natexlab}[1]{#1}
\providecommand{\url}[1]{{#1}}
\providecommand{\urlprefix}{URL }
\expandafter\ifx\csname urlstyle\endcsname\relax
  \providecommand{\doi}[1]{DOI~\discretionary{}{}{}#1}\else
  \providecommand{\doi}{DOI~\discretionary{}{}{}\begingroup
  \urlstyle{rm}\Url}\fi
\providecommand{\eprint}[2][]{\url{#2}}

\bibitem[{Avcibas et~al(2004)Avcibas, Bayram, Memon, Ramkumar, and
  Sankur}]{avcibas2004classifier}
Avcibas I, Bayram S, Memon N, Ramkumar M, Sankur B (2004) A classifier design
  for detecting image manipulations. In: Image Processing, 2004. ICIP'04. 2004
  International Conference on, IEEE, vol~4, pp 2645--2648

\bibitem[{Barni and Costanzo(2012)}]{barni2012fuzzy}
Barni M, Costanzo A (2012) A fuzzy approach to deal with uncertainty in image
  forensics. Signal Processing: Image Communication

\bibitem[{Birajdar and Mankar(2013)}]{birajdar2013digital}
Birajdar GK, Mankar VH (2013) Digital image forgery detection using passive
  techniques: A survey. Digital Investigation

\bibitem[{Castillo et~al(2007)Castillo, Melin, Kacprzyk, and Pedrycz}]{4403084}
Castillo O, Melin P, Kacprzyk J, Pedrycz W (2007) Type-2 fuzzy logic: Theory
  and applications. In: Granular Computing, 2007. GRC 2007. IEEE International
  Conference on, pp 145--145, \doi{10.1109/GrC.2007.118}

\bibitem[{Chang and Lin(2011)}]{chang2011libsvm}
Chang CC, Lin CJ (2011) Libsvm: a library for support vector machines. ACM
  Transactions on Intelligent Systems and Technology (TIST) 2(3):27

\bibitem[{Chetty and Singh(2010)}]{chetty2010nonintrusive}
Chetty G, Singh M (2010) Nonintrusive image tamper detection based on fuzzy
  fusion. International Journal of Computer Science and Network Security
  10:86--90

\bibitem[{Cox et~al(2003)Cox, Miller, and Bloom}]{cox2003digital}
Cox I, Miller M, Bloom J (2003) Digital watermarking morgan kaufmann
  publishers. San Francisco, CA

\bibitem[{Dong et~al(2009)Dong, Wang, Tan, and Shi}]{dong2009run}
Dong J, Wang W, Tan T, Shi YQ (2009) Run-length and edge statistics based
  approach for image splicing detection. In: Digital Watermarking, Springer, pp
  76--87

\bibitem[{Farid(1999)}]{farid1999detecting}
Farid H (1999) Detecting digital forgeries using bispectral analysis

\bibitem[{Farid(2009)}]{farid2009exposing}
Farid H (2009) Exposing digital forgeries from jpeg ghosts. Information
  Forensics and Security, IEEE Transactions on 4(1):154--160

\bibitem[{Fu et~al(2006)Fu, Shi, and Su}]{fu2006detection}
Fu D, Shi YQ, Su W (2006) Detection of image splicing based on hilbert-huang
  transform and moments of characteristic functions with wavelet decomposition.
  In: Digital Watermarking, Springer, pp 177--187

\bibitem[{He et~al(2006)He, Lin, Wang, and Tang}]{he2006detecting}
He J, Lin Z, Wang L, Tang X (2006) Detecting doctored jpeg images via dct
  coefficient analysis. In: Computer Vision--ECCV 2006, Springer, pp 423--435

\bibitem[{Hsiao and Pei(2005)}]{hsiao2005detecting}
Hsiao DY, Pei SC (2005) Detecting digital tampering by blur estimation. In:
  Systematic Approaches to Digital Forensic Engineering, 2005. First
  International Workshop on, IEEE, pp 264--278

\bibitem[{Johnson and Farid(2005)}]{Johnson2005}
Johnson MK, Farid H (2005) Exposing digital forgeries by detecting
  inconsistencies in lighting. In: Proceedings of the 7th workshop on
  Multimedia and security, ACM, pp 1--10

\bibitem[{Johnson and Farid(2007)}]{johnson2007exposingeye}
Johnson MK, Farid H (2007) Exposing digital forgeries through specular
  highlights on the eye. In: Information Hiding, Springer, pp 311--325

\bibitem[{Katzenbeisser et~al(2000)Katzenbeisser, Petitcolas
  et~al}]{katzenbeisser2000information}
Katzenbeisser S, Petitcolas FA, et~al (2000) Information hiding techniques for
  steganography and digital watermarking, vol 316. Artech house Norwood

\bibitem[{Kirchner and B{\"o}hme(2007)}]{kirchner2007tamper}
Kirchner M, B{\"o}hme R (2007) Tamper hiding: Defeating image forensics. In:
  Information Hiding, Springer, pp 326--341

\bibitem[{Kirchner and Bohme(2008)}]{kirchner2008hiding}
Kirchner M, Bohme R (2008) Hiding traces of resampling in digital images.
  Information Forensics and Security, IEEE Transactions on 3(4):582--592

\bibitem[{Lin et~al(2007)Lin, Lin, and Weng}]{lin2007note}
Lin HT, Lin CJ, Weng RC (2007) A note on platt's probabilistic outputs for
  support vector machines. Machine learning 68(3):267--276

\bibitem[{Majid~Valiollahzadeh et~al(2008)Majid~Valiollahzadeh, Sayadiyan, and
  Nazari}]{majid2008face}
Majid~Valiollahzadeh S, Sayadiyan A, Nazari M (2008) Face detection using
  adaboosted svm-based component classifier

\bibitem[{Ms.~Sushama(2014)}]{Rasse2014}
Ms~Sushama GR (2014) Review of detection of digital image splicing forgeries
  with illumination color estimation. International Journal of Emerging
  Research in Management \&Technology 3(3)

\bibitem[{Ng et~al(2004)Ng, Chang, and Sun}]{Ng2004}
Ng TT, Chang SF, Sun Q (2004) A data set of authentic and spliced image blocks.
  Columbia University, ADVENT Technical Report pp 203--2004

\bibitem[{Platt et~al(1999)}]{platt1999probabilistic}
Platt J, et~al (1999) Probabilistic outputs for support vector machines and
  comparisons to regularized likelihood methods. Advances in large margin
  classifiers 10(3):61--74

\bibitem[{Redi et~al(2011)Redi, Taktak, and Dugelay}]{redi2011digital}
Redi JA, Taktak W, Dugelay JL (2011) Digital image forensics: a booklet for
  beginners. Multimedia Tools and Applications 51(1):133--162

\bibitem[{Tieu and Viola(2004)}]{tieu2004boosting}
Tieu K, Viola P (2004) Boosting image retrieval. International Journal of
  Computer Vision 56(1-2):17--36

\bibitem[{Turevskiy(2014)}]{fzlmtb}
Turevskiy A (2014) Design and simulate fuzzy logic systems.
  \urlprefix\url{http://www.mathworks.com/products/fuzzy-logic/}

\bibitem[{Wang et~al(2009)Wang, Dong, and Tan}]{wang2009effective}
Wang W, Dong J, Tan T (2009) Effective image splicing detection based on image
  chroma. In: Image Processing (ICIP), 2009 16th IEEE International Conference
  on, IEEE, pp 1257--1260

\bibitem[{Yu-Feng~Hsu(2006)}]{DVMMDataset}
Yu-Feng~Hsu SF (2006) Columbia image splicing detection evaluation dataset.

\bibitem[{Zhang et~al(2009)Zhang, Cao, Zhang, Zhu, and Wang}]{Zhang2009}
Zhang W, Cao X, Zhang J, Zhu J, Wang P (2009) Detecting photographic composites
  using shadows. In: Multimedia and Expo, 2009. ICME 2009. IEEE International
  Conference on, IEEE, pp 1042--1045

\bibitem[{Zhang et~al(2008)Zhang, Zhou, Kang, and Ren}]{zhang2008study}
Zhang Z, Zhou Y, Kang J, Ren Y (2008) Study of image splicing detection. In:
  Advanced Intelligent Computing Theories and Applications. With Aspects of
  Theoretical and Methodological Issues, Springer, pp 1103--1110

\end{thebibliography}

\end{document}